\documentclass[10pt,twocolumn,letterpaper]{article}

\usepackage{iccv}
\usepackage{times}
\usepackage{epsfig}
\usepackage{graphicx}
\usepackage{amsmath}
\usepackage{amssymb}

\usepackage{multirow}

\newcommand{\tabincell}[2]{\begin{tabular}{@{}#1@{}}#2\end{tabular}}

\usepackage[pagebackref=true,breaklinks=true,letterpaper=true,colorlinks,bookmarks=false]{hyperref}

\iccvfinalcopy 

\def\iccvPaperID{1133} 

\ificcvfinal

\begin{document}

\title{Unsupervised Extraction of Video Highlights Via Robust Recurrent Auto-encoders}

\author{Huan Yang$^1$\thanks{This work was done when Huan Yang was an intern at Microsoft Research }\textbf{} \and Baoyuan Wang$^2$ \and Stephen Lin$^3$
\and David Wipf$^3$ \and Minyi Guo$^1$ \and Baining Guo$^3$\\
$^1$Shanghai Jiao Tong University \;\;\;\; $^2$Microsoft Technology \& Research \;\;\;\; $^3$Microsoft Research
}

\maketitle
\thispagestyle{empty}

\begin{abstract}
With the growing popularity of short-form video sharing platforms such as \em{Instagram} and \em{Vine}, there has been an increasing need for techniques that automatically extract highlights from video. Whereas prior works have approached this problem with heuristic rules or supervised learning, we present an unsupervised learning approach that takes advantage of the abundance of user-edited videos on social media websites such as YouTube. Based on the idea that the most significant sub-events within a video class are commonly present among edited videos while less interesting ones appear less frequently, we identify the significant sub-events via a robust recurrent auto-encoder trained on a collection of user-edited videos queried for each particular class of interest. The auto-encoder is trained using a proposed shrinking exponential loss function that makes it robust to noise in the web-crawled training data, and is configured with bidirectional long short term memory (LSTM)~\cite{LSTM:97} cells to better model the temporal structure of highlight segments. Different from supervised techniques, our method can infer highlights using only a set of downloaded edited videos, without also needing their pre-edited counterparts which are rarely available online. Extensive experiments indicate the promise of our proposed solution in this challenging unsupervised setting.
\end{abstract}

\section{Introduction}

\begin{figure}[]
\centering
{
\includegraphics[width=.99\linewidth]{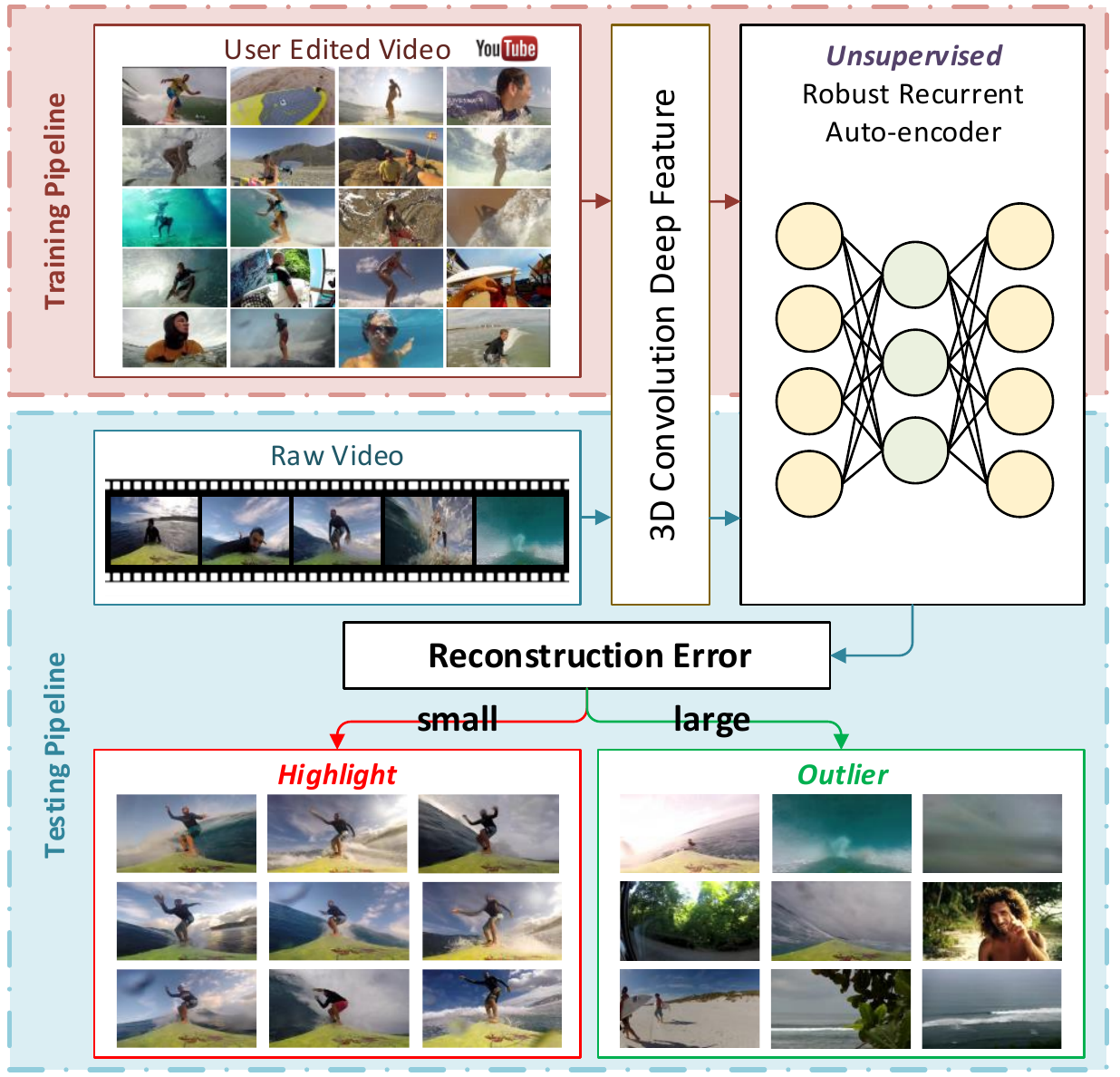}
}
\caption{Overall system pipeline}
\label{fig.pipeline}
\end{figure}
Short-form video has become a popular way for users to share their experiences on social media platforms such as {\em YouTube} and {\em Facebook}. With a well-crafted video, the user's experience can be quickly conveyed without testing the attention span of viewers. However, manually producing a highlight clip from a lengthy video, such as those captured with wearable devices like {\em GoPro} cameras, can be a time-consuming and laborious task, especially on small form-factor display devices such as smart phones. An automated tool for generating highlight clips is thus immensely desirable such that the user need only to deal with content capture.


Previous techniques address this problem either by limiting the scope to a particular context or through the use of supervised learning. The first type of method generally employs heuristic rules designed for a certain type of video, such as for broadcast sports~\cite{Rui:2000:AEH:, Otsuka:2005:HSD:,Tong:2005:HRS:,Zhang:2006:CSL:}. Though effective for their targeted settings, these techniques may not generalize well to generic, unstructured videos. In contrast, methods based on supervised learning rely on pairs of edited and raw source videos~\cite{Sun:ECCV:14} to infer highlight actions. Collecting such video pairs, however, can be a challenge. Although there exists a considerable amount of video data on the web, users typically do not upload both the raw and edited versions of a video.


In this work, we propose an {\em unsupervised} approach for generating highlight clips, using only edited videos. On the web, there are seemingly countless short-form videos that have been edited by thousands of users to contain mainly highlight sub-events. Our method capitalizes on this wealth of data by web crawling for videos in a given domain (e.g., ``surfing''), and modeling the highlights from them by inferring their common features. In this way, their raw counterparts are not needed, making it easy to scale up and collect more training data. Additionally, since the videos have been edited by a large community of users, the risk of building a biased highlight model is greatly reduced in comparison to using a training set constructed from a small number of users.

There exist significant challenges with this approach: (1) Although most people have a common notion of what the highlights should be in a certain video domain such as ``surfing'', there nevertheless may exist subjective differences among users (e.g., whether entering the water is highlight-worthy). (2) A query on a given keyword, such as ``GoPro surfing'', may return some noisy results that are not relevant to the targeted domain. (3) No information other than the queried videos themselves are available to be leveraged. Unlike in previous supervised learning approaches~\cite{Sun:ECCV:14, DBLP:conf/eccv/PotapovDHS14, GygliECCV14,DPP:NIPS2014}, there are no unedited counterpart videos that can be used to identify what is and is not important to keep in a highlight clip.

To address these issues, we propose to identify and model highlights as the most common sub-events among the queried videos and remove uninteresting or idiosyncratic snippet selections that occur relatively infrequently. Our method accomplishes this via an auto-encoding recursive neural network (RNN)~\cite{Vincent:2008:ECR} that is trained from positive examples to reconstruct highlight input instances accurately while non-highlights are not. Intuitively, since highlights are assumed to occur much more frequently among the queried videos, they will have clustered distributions in the feature space while non-highlights occur as outliers.

We formulate the auto-encoder with two main features. Since training data crawled from the web is generally noisy (containing some amount of negative examples), we propose a novel shrinking exponential loss function that makes the auto-encoder training robust to noisy data. With the shrinking exponential loss, outliers are gradually identified in the training data and their influence in the auto-encoder training is progressively reduced. The other main feature accounts for the temporal structure of video highlights (e.g., standing up on the surfboard, riding the wave, and then falling into the ocean). To take advantage of this contextual dependency, we construct the auto-encoder with bidirectional long short term memory (LSTM)~\cite{LSTM:97} cells, which have been shown in areas such as speech recognition~\cite{LSTM:SPEECH:} to effectively model long-range context in time-series data.


The main technical contributions of this work are the formulation of video highlight detection as an unsupervised learning problem that takes advantage of the abundance of short-form video on the web, and modeling video highlight structure through a robust recurrent auto-encoder with a shrinking exponential loss function and bidirectional LSTM cells. With the proposed unsupervised technique, we show promising results that approach the quality of supervised learning but without the burden of collecting pre- and post-edit video pairs.


\section{Related Work}
As defined in \cite{Sun:ECCV:14}, a video highlight is a moment of major or special interest in a video. Generating highlight clips is thus different from the task of video summarization, which instead accounts for factors such as ``diversity'' and ``representativeness'' to convey a brief but comprehensive synopsis of a video. Despite this different goal, we review methods for video summarization in addition to video highlight detection because of similarities between the two topics.

\paragraph{Video Highlight Detection}
Traditionally, video highlight detection has primarily been focused on broadcast sports videos \cite{Rui:2000:AEH:, Otsuka:2005:HSD:, Tong:2005:HRS:, Zhang:2006:CSL:}. These techniques usually employ features that are specific to a given sport and the structure of sports broadcasts, and are therefore hard to generalize to the more generic videos of ordinary users. Recently, the scope of highlight detection was expanded to a broad range of videos in \cite{Sun:ECCV:14}, where a latent SVM model was proposed to rank highlight segments ahead of less interesting parts through supervised learning. Good results have been demonstrated with this approach on action videos such as those captured by GoPro cameras. However, the supervised learning requires each training example to be a video pair composed of an edited video and its corresponding raw source video. Such training pairs are difficult to collect in large quantities, since users tend to upload/share only the edited versions, as the source videos generally are too long for general consumption. This makes large scale training mostly impractical for this approach. In addition, it employs a computationally-intensive feature representation, namely dense trajectories \cite{Heng13_IJCV_MBH}, which involves computing dense optical flows and extracting low-level features such as HoG, HoF and MBH prior to Fisher vector encoding. By contrast, our method performs unsupervised learning from edited videos only, and utilizes generic deep learning features which are more computationally efficient and more accurate in characterizing both appearance and motion.

\paragraph{Video Summarization} A comprehensive review of video summarization can be found in \cite{Truong:2007:VAS}. Among recent methods, several are guided by saliency-based properties such as attention~\cite{AttentionModel:05}, interestingness~\cite{Ngo:2005:VSS,Kang:2006:SVM,GygliECCV14}, and important people and objects~\cite{Eogcentric:12}. However, the most salient frames in a video do not necessarily correspond to its highlights, which depend heavily on the video domain. Others aim to provide a comprehensive synopsis based on connectivity of sub-events~\cite{Lu:Stroy_driven} or diversity of video segments~\cite{cvpr/ZhaoX14a,Liu:2002:OAS:}. While this helps to provide a complete overview of a video, a highlight clip instead is focused on only certain segments that are specific to the video domain, while discarding the rest.

Video summarization techniques have also employed supervised learning. Methods along this direction use category-specific classifiers for importance scoring~\cite{DBLP:conf/eccv/PotapovDHS14} or learn how to select informative and diverse video subsets from human-created summaries~\cite{DPP:NIPS2014}. These supervised techniques have led to state-of-the-art video summarization results, but are not suitable for highlight clip generation. In addition, it is generally feasible to have only a limited number of users to annotate training videos, which may lead to a biased summarization model. Our method instead learns from videos pooled from many users on the web to obtain a highlight model less influenced by the particular preferences of certain people.

\paragraph{Novelty Detection} The one-class learning of our recurrent auto-encoder is related to works on novelty detection, which aim to identify outliers from an observed class. In \cite{Marchi2015}, novelty detection is performed for audio features using an auto-encoder with LSTM. Our concurrent work deals instead with RGB video data, for which meaningful features are more challenging to extract. We address this through temporal video segmentation, extraction of high-level spatial-temporal features for each segment, and then temporal pooling, before feeding to the auto-encoder. Moreover, we introduce a novel shrinking loss function to address the noisy training data in our context.

There exist other unsupervised novelty detection techniques that could potentially be applied to our problem, such as the unsupervised one-class learning in \cite{WeiLiu2014} or outlier-robust PCA~\cite{Xu2013}. In our work, we chose the auto-encoder as our basic unsupervised framework because its properties are well-suited to our application, such as scalability, easy parallelization, and seamless integration with LSTM cells. How to customize other novelty detection techniques for video highlight detection is a potential direction for future investigation.

\begin{figure}[t!]
\centering
{
\includegraphics[width=.99\linewidth]{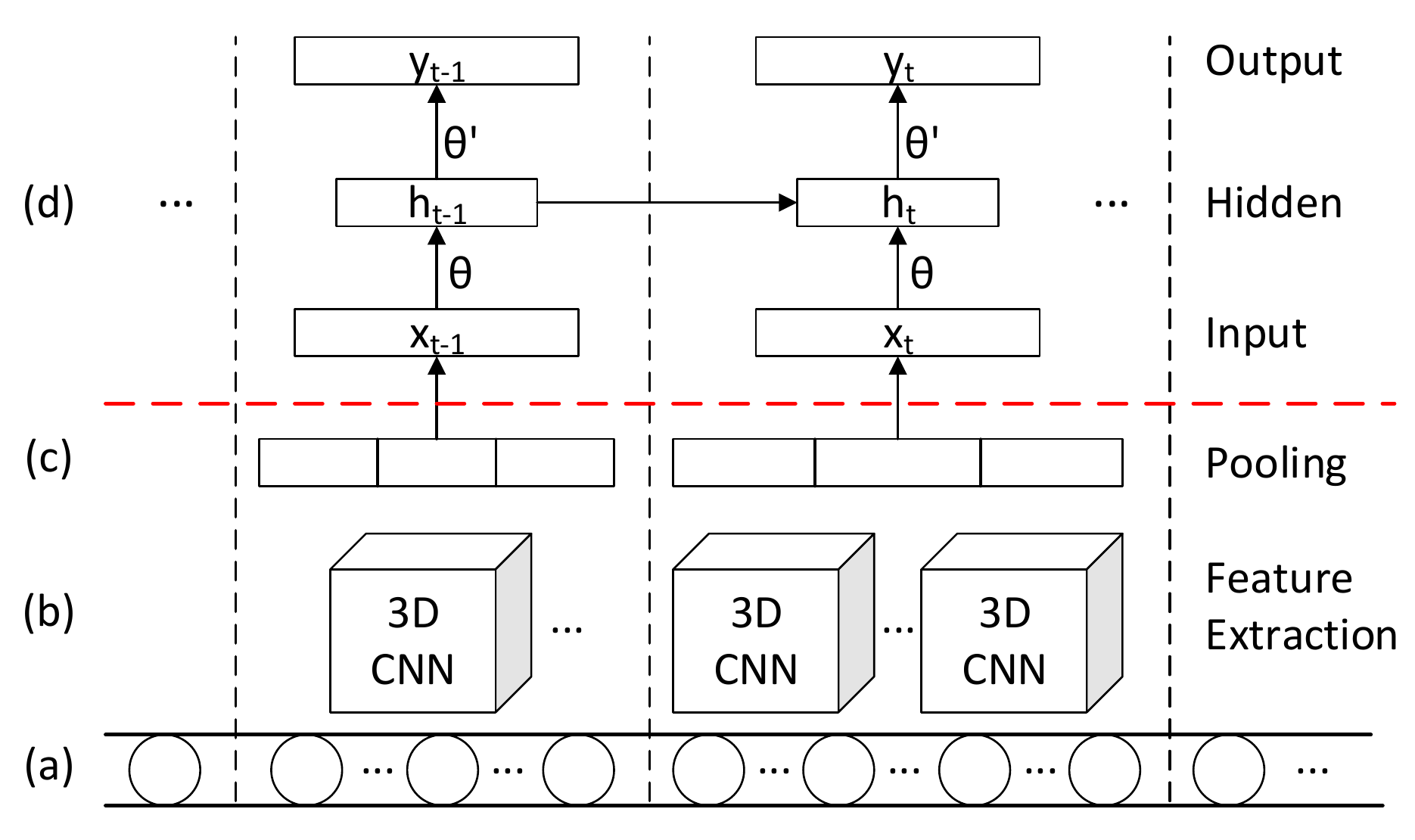}
}
\caption{Conceptual illustration of our overall pipeline and architecture. (a) Each video is first segmented into multiple short snippets. (b) Then we apply a pre-trained 3D convolution neural network model \cite{C3D:14} to extract spatial-temporal features. (c) This is followed by a temporal pooling scheme that respects the local ordering structure within each snippet. (d) The robust recurrent auto-encoder with the proposed companion loss is then employed to capture the long range contextual structure.}
\label{fig.CNN_RNN}
\end{figure}

\section{Auto-Encoder-Based Removal of Outliers}
An auto-encoder is an artificial neural network~\cite{Rumelhart:1986:LIR:} that is trained to reconstruct its own input. A common use of auto-encoders is for dimensionality reduction, where if the hidden layers have fewer nodes than the input/output layers, then the activations of the final hidden layer can be taken as a compressed representation of the original signal. An auto-encoder operates by taking an input vector $x \in [0,1]^d$ and first mapping it to a hidden layer $h \in [0,1]^{d'}$ through a deterministic function $f_{\theta}(x) = s(Wx + b) $, parameterized by $\theta = \{W,b\}$ where $W$ is a $d' \times d$ weight matrix, $b$ is a bias vector, and $s$ is the activation function, such as a sigmoid or rectified linear unit (ReLU). This hidden layer is then mapped to an output layer $y \in [0,1]^d$ with the same number of nodes as the input layer through another function $g_{\theta'}(x) = s(W'x+b')$, with $s$ being a linear function at this layer. Thus, $y=g_{\theta'}(f_{\theta}(x))$. Backpropagation with stochastic gradient descent (SGD) is employed to optimize the parameters $\theta$ and $\theta'$ via the following loss function
\begin{eqnarray}
(\theta^*, \theta'^*) &=& \arg\min_{\theta, \theta'} \frac{1}{n} \sum_{i=1}^{n} L(x^i,y^i) \\
                      &=&  \arg\min_{\theta, \theta'} \frac{1}{n} \sum_{i=1}^{n} L(x^i,g_{\theta'}(f_{\theta}(x^i))),
\end{eqnarray}
where $L$ is generally defined as the squared error $L(x,y) = \|x - y\|^2$ and each $x^i$ is a training sample. When $d' < d$, the auto-encoder acts as a compression neural network that works surprisingly well for single-class document classification \cite{Manevitz:2002:OSD:} and novelty detection \cite{Japkowicz95anovelty}. The key idea is that inlier (or positive) instances are expected to be faithfully reconstructed by the auto-encoder, while outliers (or negative) instances are not. So one can classify an unseen instance by checking its reconstruction error from the auto-encoder. Our work is partially inspired by the applications of auto-encoders for novelty detection, and we present two significant modifications to tailor them for our highlight detection problem.

\section{Our Approach}
In this section, we introduce a new domain-specific video highlight system. Our core idea is to leverage the wealth of crowd-sourced video data from the web and automatically learn a parametric highlight detection model. Before describing the technical details, we first introduce the overall system pipeline, illustrated in Fig.~\ref{fig.pipeline}.

\subsection{Overview} \label{sec:overview}
\paragraph{Acquisition of Training Data:} Our system starts with a simple video crawling step. Given the keyword for a specific domain, such as ``GoPro Surfing", our system automatically retrieves a large number of videos from YouTube with this keyword. We restrict this search to only short-form videos (i.e., videos less than four minutes long), since such videos from social media are likely to have been edited by the user. With this approach, we can easily build a large-scale training set edited by many different people. Since our system learns what is a highlight based on commonalities among different videos, having a diverse user pool helps to avoid biases in this inference. Let us denote the training video set as $S=\{v_1,v_2,...,v_N\}$. Our system then automatically models the highlights in $S$ through the use of a proposed auto-encoder.

\paragraph{Temporal Segmentation:}
A highlight can be defined as a motion or moment of interest with respect to the video domain context. So we first segment each video $v_i, i\in[1,N]$ into multiple non-uniform snippets using an existing temporal segmentation algorithm~\cite{DBLP:conf/eccv/PotapovDHS14}. We added a constraint to the segmentation algorithm to ensure that the number of frames within each snippet lies in the range of $[48,96]$. The segmented snippets serve as the basic units for feature extraction and subsequent learning and inference (Fig.~\ref{fig.CNN_RNN}(a)). After segmentation, a highlight sequence in the edited video might correspond to one or multiple consecutive snippets. At runtime, our system outputs the highlight confidence score for each snippet within the input video.

\paragraph{Feature Representation:}
Recent work in deep learning has revealed that features extracted at higher layers of a convolutional neural network are generic features that have good transfer learning capabilities across different domains~\cite{Decaf,GigaSUN, DeepVideo14,Two_Stream:Action:NIPS2014,C3D:14}. An advantage of using deep learning features is that there exist accurate, large-scale datasets such as Places \cite{GigaSUN} and One-million Sports \cite{DeepVideo14} from which they can be extracted. In addition, GPU-based extraction of such features is much faster than that for the traditional bag-of-words and Fisher vector models. For example, C3D features~\cite{C3D:14} are $50\times$ faster to extract than dense trajectories~\cite{Heng13_IJCV_MBH}. We therefore extract C3D features, by taking sets of $16$ input frames, applying 3D convolutional filters, and extracting the responses at layer ``FC6" as suggested in~\cite{C3D:14} (Fig.~\ref{fig.CNN_RNN}(b)). This is followed by a temporal mean pooling scheme to maintain the local ordering structure within a snippet (Fig.~\ref{fig.CNN_RNN}(c)). Then the pooling result serves as the final input feature vector to be fed into the auto-encoder (Fig.~\ref{fig.CNN_RNN}(d)).

\paragraph{Unsupervised Learning:} After building the representation for each snippet, we learn a discriminative model for highlight detection using a novel robust recurrent auto-encoder. It is assumed that the collected training set $S$ for each domain contains coherence in the highlights so that they can be distinguished from the remaining parts by reconstruction error. We note that this cannot be treated as a multiple instance learning problem. Since a video does not necessarily contain at least one highlight snippet, such as when the video is actually unrelated to the keyword, the bag and instance relationship is hard to define.

\begin{figure}[t!]
\centering
{
\includegraphics[width=.75\linewidth]{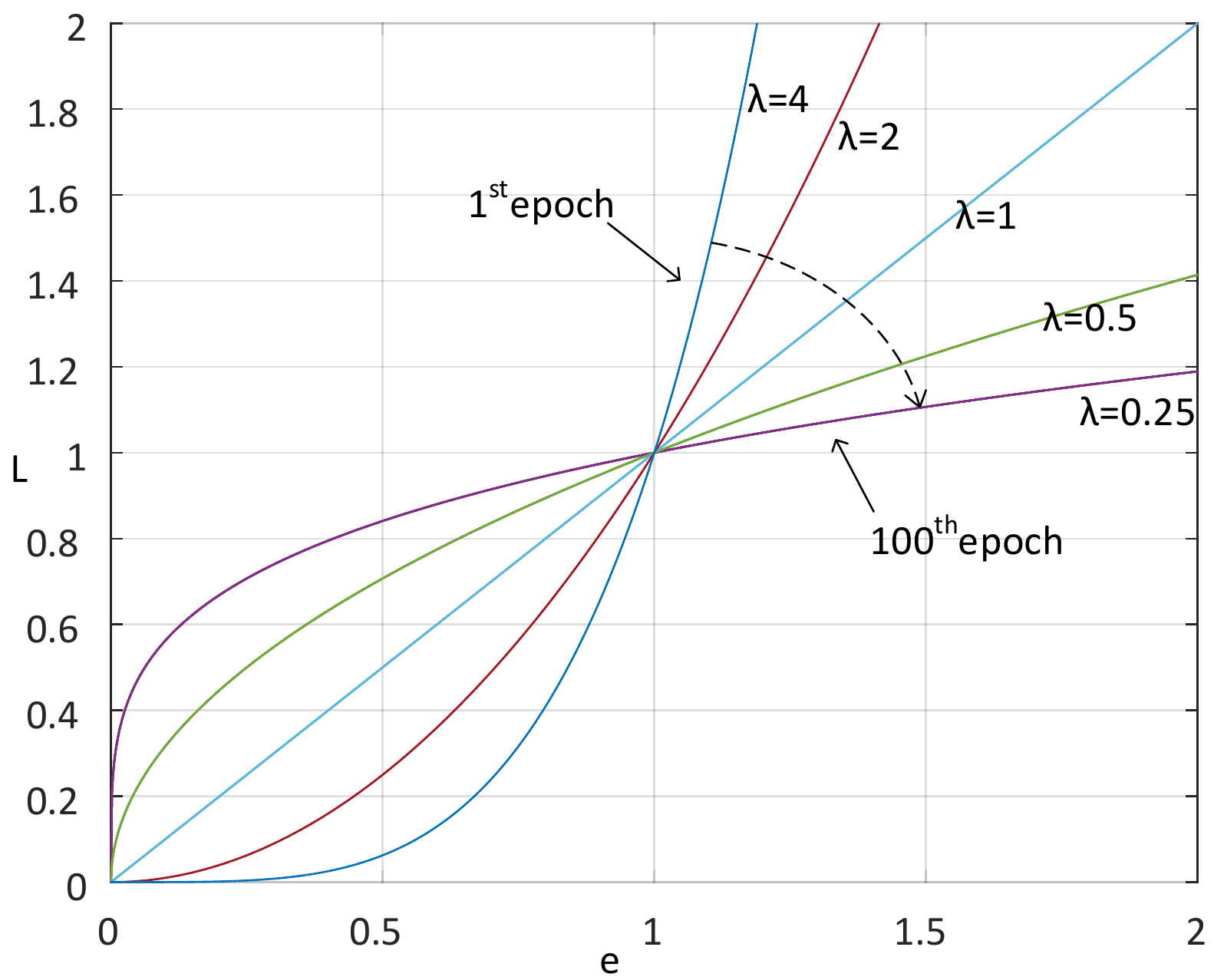}
}
\caption{Unlike the squared loss used in standard auto-encoders, we propose a more general exponential loss $L=e^\lambda$ with its exponential parameter $\lambda$ shrinking during the course of training. The horizontal axis $e$ represents the reconstruction error, while the vertical axis $L$ signifies the loss.}
\label{fig.RobustAE}
\end{figure}

\subsection{Robust Auto-encoder Via Shrinking Exponential Loss}
\label{sec:shrinkingLoss}
In training an auto-encoder, it is assumed that the training data consists only of positive instances, which the auto-encoder learns to replicate as output. However, since we obtain our training data through web crawling, we cannot guarantee that the data is free of negative instances. To train an auto-encoder that is robust to such noise, we propose a shrinking exponential loss function that helps to reduce the influence of negative examples.  This function is defined by
\begin{eqnarray}\label{equ:loss}
L(x,y)  &=& (\parallel x - y \parallel_2 ^2)^\lambda \\
\lambda &=& f(epo),
\end{eqnarray}
where $\lambda$ is a function $f$ of the current epoch number $epo$, and $L$ is equivalent to the standard squared loss when $\lambda=1$. With backpropagation for network training, an example that has a large loss gradient will contribute more to the training than other examples whose corresponding loss gradient is small. Since network parameters are randomly initialized, all the examples will generally have a large loss at the beginning. So to expedite convergence in the early stages of training, we utilize a relatively large value of $\lambda$, which magnifies the loss gradients. As the positive examples share commonalities and are assumed to be more clustered relative to the negative examples, the network parameters will start to converge in a manner such that the positive examples become more accurately reconstructed. At the same time, it is desirable to shrink the exponent $\lambda$ in order to decrease the influence of negative examples, which on average have larger loss gradients because of their more dispersive nature.

To accomplish this we define $f$ to be monotonically decreasing with respect to $epo$ as shown in Figure~\ref{fig.RobustAE}, with values greater than $1$ in early stages to promote convergence, and shrinking to less than $1$ in later stages to reduce the impact of outliers with higher reconstruction error. In our work, we empirically define $f$ such that $\lambda$ varies linearly in the range of $[e,s]$, with $e \in(0,1]$ and $s \geq 1$, giving
\vspace{-1mm}
\begin{eqnarray}
f(epo) = s - \frac{epo * (s-e)}{\Gamma},
\end{eqnarray}
where $\Gamma$ is the total number of training epochs.  As demonstrated later in our experiments, this formulation of our shrinking exponential loss provides greater robustness to negative examples than the standard squared loss for which $\lambda$ is fixed to $1$.

\subsection{Recurrent Auto-Encoder with LSTM Cells}
Each highlight snippet has a certain degree of dependence on its preceding and even subsequent frames. Taking surfing as an example, a surfer must first stand up on the board before riding a wave. The ``stand up" action thus provides contextual information that can help to discriminate subsequent surfing highlights. In this work, we take advantage of such temporal dependencies through the use of long short term memory cells.

Given an input sequence $x=(x_1,x_2,...,x_T), x_t \in R^d, t\in[1,T]$, a recurrent neural network (RNN) designed as an auto-encoder needs to first compute a hidden vector sequence $h=(h_1,h_2,...,h_T), h_t \in R^{d'}, d' < d$ such that it outputs a reconstructed sequence $y=(y_1,y_2,...,y_T)$ where $y_t\approx x_t$. This can be solved through iterations of the following equations:
\begin{eqnarray}
h_t &=& \mathcal{H}(W_{ih}x_t + W_{hh}h_{t-1} + b_h)\\
y_t &=& W_{ho}h_t + b_o
\end{eqnarray}
where $W_{ih}$ and $W_{hh}$ denote the input-hidden and hidden-hidden weighting matrices, and $b_h$ and $b_o$ represent bias vectors. $\mathcal{H}$ is the hidden layer activation function, usually chosen as an element-wise sigmoid.

\begin{figure}[t]
\centering
{
\includegraphics[width=.9\linewidth]{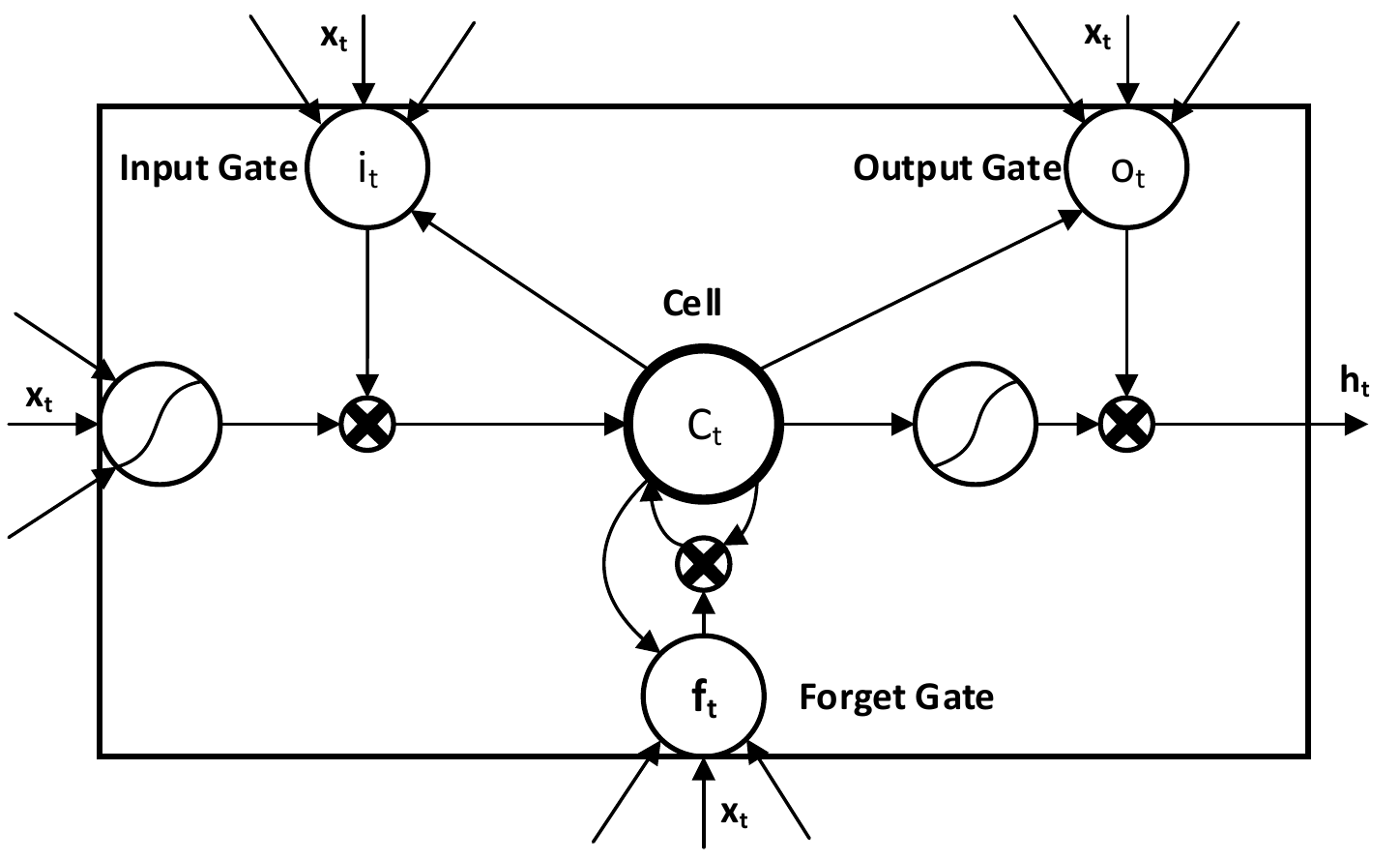}
}
\caption{Long short term memory cell (regenerated from \cite{LSTM:SPEECH:}).}
\label{fig.LSTM_Cell}
\end{figure}

For time-series data, it has been found that LSTM cells~\cite{LSTM:97} are more effective at finding and modeling long-range context along a sequence, as shown in recent works on speech recognition~\cite{LSTM:SPEECH:} and human action recognition~\cite{LSTM:Video}. Figure~\ref{fig.LSTM_Cell} shows a typical structure of a LSTM cell, which operates by learning gate functions that determine whether an input is significant enough to remember, whether it should be forgotten, and when it should be sent to output. By storing information over different time ranges in this manner, a LSTM-RNN is better able to classify time-series data than a standard RNN. Inspired by these works, we propose to integrate LSTM cells as the hidden nodes in the auto-encoder network. With LSTM cells, $\mathcal{H}$ is then defined by the following composite functions:
\begin{eqnarray}
c_t&\!=\!& f_tc_{t-1} + i_t\tanh(W_{xc}x_t + W_{hc}h_{t-1} + b_c)\\
i_t&\!=\!&\sigma(W_{xi}x_t + W_{hi}h_{t-1} + W_{ci}c_{t-1} + b_i)\\
f_t&\!=\!&\sigma(W_{xf}x_t + W_{hf}h_{t-1} + W_{cf}c_{t-1} + b_f)\\
o_t&\!=\!&\sigma(W_{xo}x_t + W_{ho}h_{t-1} + W_{co}c_t + b_o)\\
h_t&\!=\!& o_t \tanh(c_t)
\end{eqnarray}
where $\sigma$ is the logistic sigmoid function, and $i,f,o$ are respectively the \textit{input gate, forget gate and output gate}, which take scalar values between $0$ and $1$. $c$ denotes the \textit{cell} activation vectors which have the same size as the hidden vector $h$. The terms in the $W$ matrices represent the connections, for example, with $W_{xi}$ denoting the input-input gate matrix and $W_{hf}$ representing the hidden-forget gate matrix.

In practice, we use bidirectional LSTM cells to model both forward and backward dependencies. For further details on LSTM, please refer to \cite{LSTM:97} and \cite{LSTM:SPEECH:}.

\section{Experiments}
\subsection{Datasets}
\begin{table}[t]
  \centering
  \begin{tabular*}{.5\textwidth}{ c@{\extracolsep\fill} c c c c}
    \hline
    $ \tabincell{c}{} $ & \tabincell{c}{\# of train\\ videos} & \tabincell{c}{\# of test\\ videos} & \tabincell{c}{Coverage of\\  train set} & \tabincell{c}{ H-ratio of\\ train set} \\ \hline
    $freeride$  & $912$  & $27$  & $0.63$ & $0.33$    \\ \hline
    $parkour$  & $781$  & $29$  & $0.83$ & $0.32$    \\ \hline
    $skating$  & $940$  & $34$ & $0.67$ & $0.26$     \\ \hline
    $skiing$  & $945$  & $50$  & $0.83$ & $0.32$    \\ \hline
    $skydiving$  & $762$  & $30$& $0.83$ & $0.38$    \\ \hline
    $surfing$  & $928$  & $52$ & $0.80$ & $0.31$    \\ \hline
    $swimming$  & $1015$  & $29$ & $0.73$ & $0.25$    \\ \hline
  \end{tabular*}
  \caption{Statistical information on our dataset}
  \label{tab:dataStats}
\end{table}
\paragraph{Edited Videos for Training Set}
As mentioned in Sec.~\ref{sec:overview}, our system can automatically harvest domain-specific datasets from the web using keywords and other search conditions. Unlike previous data crawling systems such as in \cite{Sun:ECCV:14}, ours need only obtain short-form videos edited by users, and not their corresponding raw versions. The editing operations may have been applied either by post-processing or through selective capture. For evaluating the performance of our approach, we have crawled more than 6500 short-form videos totaling about 13800 minutes from YouTube\footnote{We use the tool ``youtube-dl" in http://rg3.github.io/youtube-dl/ to crawl for videos using the domain name and with a search condition of less than four minutes.}. The search terms include \textit{freeride, parkour, skating, skiing, skydiving, surfing and swimming}. Temporal segmentation of the videos yields $442075$ snippets. Compared with the training set used in~\cite{Sun:ECCV:14}, ours is more than $10\times$ longer, and can easily be expanded further.

To obtain a better sense of how this set correlates with the underlying highlights, we present two quantitative measurements, \textit{Coverage} and \textit{H-ratio}. \textit{Coverage} refers to the percentage of videos that contain at least one highlight snippet (the basic unit of segmentation), while \textit{H-ratio} is the percentage of highlight snippets among all the snippets within the set. We calculate these measures from a randomly selected subset of 30 videos from each domain with manual highlight annotations. The statistics of our dataset in Table~\ref{tab:dataStats} show the \textit{Coverage} value to be about $70\%$ and the \textit{H-ratio} about $30\%$ in each domain, indicating that users indeed tend to share edited videos which have a significant amount of highlight content.

\paragraph{Raw Videos for Testing Set} For each domain, we also manually collected about 30 raw videos (see third column of Table~\ref{tab:dataStats}) which do not correspond to the training videos, and asked six people to annotate the highlights for each video. A snippet is considered to be a highlight only if they were labeled as such by at least four of the people. Annotations were collected by simply having the users mark each snippet as highlight or not. The total length of the testing videos is about 700 minutes, which is 2.5x longer than the testing data used by \cite{Sun:ECCV:14}.

\subsection{Implementation \& Evaluation Details}
Our system was implemented in C++ and runs on a workstation with a 3.1GHz dual-core CPU and an Nvidia Tesla K40 graphics card. For all the training and testing videos, we first segment them into multiple snippets using the method described in Sec.~\ref{sec:overview}. For the standard auto-encoder, we treat each snippet as one example, while for the bidirectional recurrent auto-encoder with LSTM cells, we treat the nine-snippet sequence centered on the current snippet as one example. We found that increasing the sequence length has little effect on the performance. For all the experiments conducted in this paper, we use basic auto-encoders with only one hidden layer and linear activation functions. The number of hidden nodes was chosen to be half that of the input layer. Raw features are extracted from the ``FC6" layer of the C3D \cite{C3D:14} network prior to PCA dimensionality reduction, which maintains $90\%$ of the total energy. We then apply simple mean pooling within each snippet, which provides performance for C3D similar to that of segmented temporal pooling. The performance benefits of employing the 3D spatial-temporal C3D features rather than the 2D single-frame features of Places CNN is shown in Table~\ref{tab:2D3D}, where the results are obtained using standard auto-encoders trained on each type of feature without applying PCA.

\begin{table}[t]
  \centering
  \begin{tabular*}{.4\textwidth}{ c@{\extracolsep\fill} c c }
    \hline
     & Places CNN \cite{GygliECCV14} & C3D \cite{C3D:14} \\ \hline
    $freeride$   & $0.241$  & $0.302$    \\
    $parkour$    & $0.323$  & $0.425$    \\
    $skating$    & $0.310$  & $0.304$    \\
    $skiing$     & $0.462$  & $0.388$    \\
    $skydriving$ & $0.337$  & $0.433$    \\
    $surfing$    & $0.501$  & $0.539$    \\
    $swimming$   & $0.350$  & $0.320$    \\ \hline
    Overall mAP  & $0.361$  & $0.387$    \\ \hline
  \end{tabular*}
  \caption{Comparison of mean average precision (mAP) between 2D and 3D CNN features with a standard auto-encoder. Both types of features are 4096-D vectors. The Places CNN features were temporally pooled by dividing each snippet into two uniform sub-snippets and performing mean pooling on each, while the C3D features were simply mean pooled within each whole snippet.}
  \label{tab:2D3D}
\end{table}

Our results can be reproduced through the following network training parameters which were set without careful tweaking: learning rate of $0.01$, weight decay of $0.0005$, and momentum of $0.9$. We set the maximum epoch number ($\Gamma$) to 100, and let $\lambda$ shrink from $2$ to $0.25$.

For evaluation, we use the same metric as in \cite{Sun:ECCV:14}. For each video, we sort the snippets in ascending order based on their corresponding reconstruction errors. We then compute the hit rate of the top $K$ snippets with respect to their ground-truth. Finally this number is averaged across the entire testing set to obtain the mean average precision (mAP).

\subsection{Results and Discussion}
We compare our robust recurrent auto-encoder (RRAE) to other unsupervised alternatives and to the supervised learning technique of \cite{Sun:ECCV:14}. Before that, we examine the effect of different parameters for the shrinking exponential loss.

\subsubsection{Effect of Shrinking Exponential Loss}
As discussed in Sec.~\ref{sec:shrinkingLoss}, using a shrinking exponential loss during training helps to reduce the influence of outliers compared with the standard fixed loss. This is validated in Table~\ref{tab:overall_results} by comparing the standard auto-encoder (AE) with its robust version based on the new shrinking exponential loss (robust AE).
Intuitively, by gradually changing the shape of loss functions through shrinking $\lambda$, the gradients of non-highlight snippets become relatively small so that their influence on network training is gradually reduced. As shown in Table \ref{Tab:veryingLambdas}, shrinking $\lambda$ consistently matches or surpasses $s=e=1$ (the standard fixed squared loss). We also observed in this study that shrinking to small values (e.g., $\lambda=0.125$) may be unfavorable in some cases as this ends up ignoring too many examples, including inliers. Although carefully tweaking the shrinking range can improve results, we found that going from $\lambda>1$ to $\lambda<1$ generally works well for the domain categories we examined. Nevertheless, design of a more optimal shrinking scheme would be an interesting direction for future work.

\begin{table}[t]
\centering
\begin{tabular}{c|c|c c c}
\hline
\multirow{2}{*}{$s$} & \multirow{2}{*}{$e$} & \multicolumn{3}{c}{mAP}\\ \cline{3-5}
                       &                      & freeride & skiing   & skydiving \\\hline
\multirow{2}{*}{0.5}   & 0.25                 & 0.260 & 0.429 & 0.330  \\
                      & 0.5                  & 0.277 & 0.443 & 0.356  \\ \hline
\multirow{3}{*}{1}     & 0.25                 & 0.278 & 0.447 & 0.355  \\
                       & 0.5                  & 0.274 & 0.423 & 0.361  \\
                       & 1                    & 0.264 & 0.423 & 0.336  \\  \hline
\multirow{4}{*}{2}     & 0.25                 & 0.274 & 0.437 & 0.362  \\
                       & 0.5                  & 0.286 & 0.476 & 0.373 \\
                       & 1                    & 0.289 & 0.476 & 0.360\\
                       & 2                    & 0.283 & 0.453 & 0.366  \\ \hline
\end{tabular}
\caption{Results with different shrinking exponential parameters. $\lambda$ shrinks linearly from $s$ to $e$. Due to limited space, we only show results for three domain categories.}
\label{Tab:veryingLambdas}
\end{table}

\begin{table*}[t!]
  \centering
  \begin{tabular*}{.9\textwidth}{ c@{\extracolsep\fill}| c | c c c c c c}
    \hline
    $ \tabincell{c}{}$ & $ \tabincell{c}{LRSVM \cite{Sun:ECCV:14}} $ & \tabincell{c}{PCA} & \tabincell{c}{OCSVM} & AE &\tabincell{c}{Robust AE} & \tabincell{c}{ Recurrent AE} & RRAE \\ \hline
    $freeride$ & *& $0.235$  & $0.258$  & $0.268$ & $0.277$ & $0.277$ & $\textbf{0.288}$ \\
    $parkour$  & $0.246$& $0.377$  & $0.445$  & $0.507$ & $0.508 $& $0.618$ & $\textbf{0.675}$\\
    $skating$  & $0.330$& $0.251$  & $0.297$  & $0.308$ & $0.306$ & $0.322$ & $\textbf{0.332}$\\
    $skiing$   & $0.337$& $0.388$  & $0.412$  & $0.428$ & $0.472$ & $0.478$ & $\textbf{0.485}$\\
    $skydiving$&  *& $0.376$ & $0.332$  & $0.335$ & $0.364$ & $0.338$ & $\textbf{0.390}$\\
    $surfing$  & $0.564$& $0.525$  & $0.484$  & $0.494$ & $0.534$ & $0.565$ & $\textbf{0.582}$\\
    $swimming$ & *& $0.274$  & $0.238$  & $0.255$ & $0.277$ & $0.275$ & $\textbf{0.283}$\\ \hline
    mAP        & & $0.347$  & $0.352$  & $0.371$ & $0.391$ & $0.410$ & $\textbf{0.434}$\\ \hline

  \end{tabular*}
  \caption{Performance results of our methods and several baseline methods, all using C3D features \cite{C3D:14}. The dimensionality of C3D features is reduced from 4096 by a domain specific PCA that keeps $90\%$ of the total energy.}
  \label{tab:overall_results}
\end{table*}



\subsubsection{Unsupervised Learning Comparisons}
There exist other unsupervised learning techniques that could be applied to this problem. In addition to the standard auto-encoder (AE), two frequently used methods for anomaly detection and outlier removal are Principal Components Analysis (PCA) and One-class Support Vector Machines (OCSVM) \cite{OCSVM:2001}. For PCA, we project the original $d$ dimensional input vector into a $d'$ dimensional subspace, with $d' < d$ as in the auto-encoder. Then snippets with small PCA reconstruction error are taken as highlights. For one-class SVM, we use the LibSVM implementation where its parameters $\gamma$ (RBF kernel width) and $\nu$ (for controlling the outlier ratio) are chosen using a simple line search based on testing error. For the different domain classes, we found that the optimal $\nu$ lies in the range of $[0.5,0.9]$, while $\gamma$ lies in $[1,10]$.


Comparisons among these methods are presented in Table~\ref{tab:overall_results}. Our robust recurrent auto-encoder consistently outperforms AE, PCA and OCSVM on all the domain categories. Although OCSVM works better than PCA, it requires much more computation because of its nonlinear kernel. The table additionally includes comparisons to partial versions of our techniques, namely recurrent AE without the shrinking exponential loss, and non-recurrent AE but with the shrinking loss (denoted as Robust AE). From these results, we can see that when the standard auto-encoder is equipped with LSTM cells (recurrent AE), the performance is boosted by more than $10\%$, from $0.371$ to $0.410$. This indicates the importance of modeling the temporal contextual structure of video highlights.

\begin{figure*}[t!]
\centering
{
\includegraphics[width=0.99\linewidth]{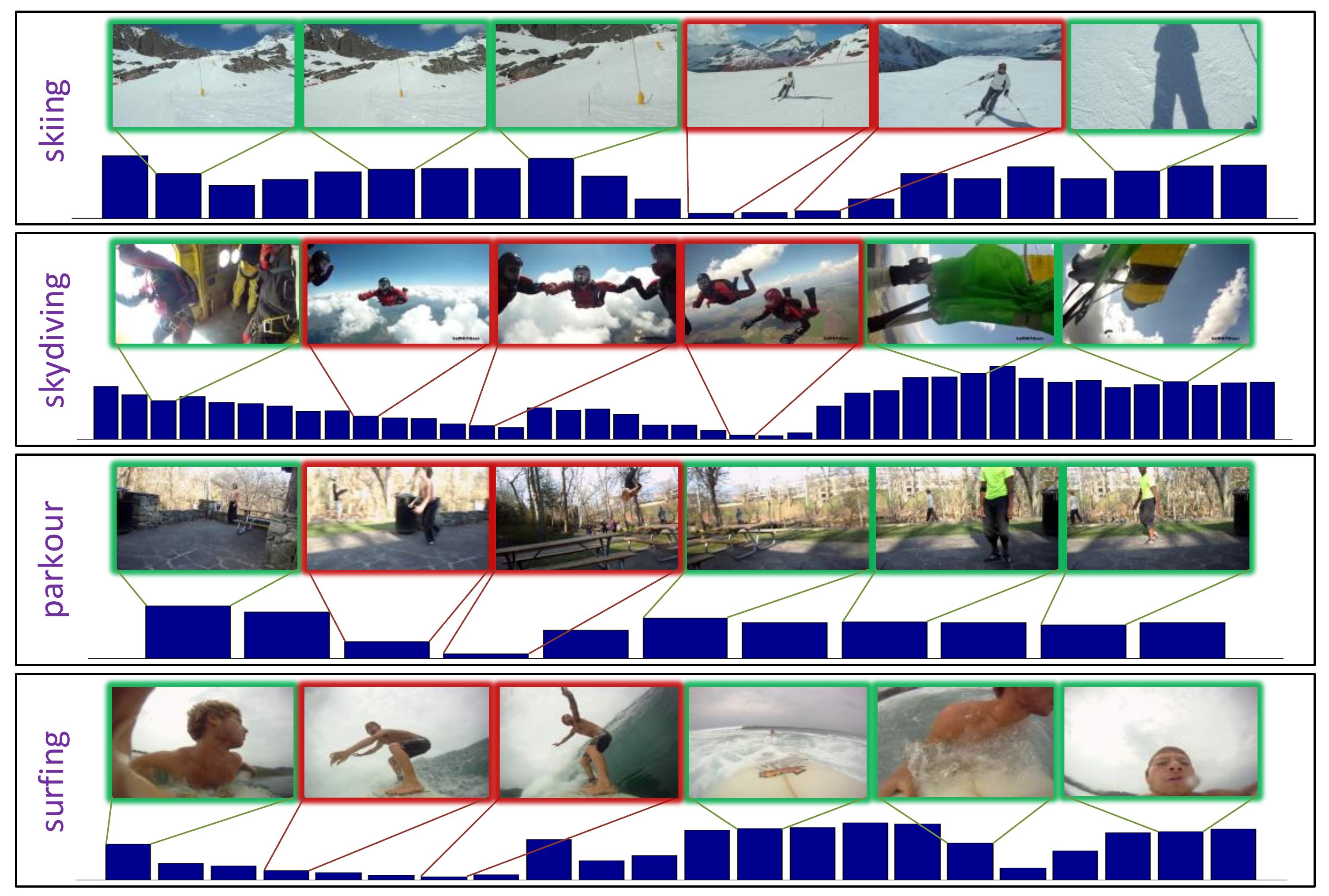}
}
\caption{Highlight detection results in different video domains. The blue bar represents reconstruction error, where highlights tend to have smaller errors than non-highlight snippets. The red borders indicate snippets detected as highlights.}
  \vspace*{-0.2cm}
\label{fig.gallery}
\end{figure*}

Some of our detection results in different video domains are illustrated in Figure~\ref{fig.gallery}. The blue bars represent reconstruction error, with smaller values having a higher probability of being a highlight snippet.

\begin{table}[t]
  \centering
  \begin{tabular*}{.4\textwidth}{ c@{\extracolsep\fill} c c}
    \hline
    $$ & $Supervised \cite{Sun:ECCV:14}$   & $RRAE$ \\  \hline
    $dog$  & $0.60$  & $0.49$      \\
    $gymnastics$  & $0.41$  & $0.35$      \\
    $parkour$  & $0.61$  & $0.50$      \\
    $skating$  & $0.62$  & $0.25$      \\
    $skiing$  & $0.36$  & $0.22$      \\
    $surfing$  & $0.61$  & $0.49$      \\ \hline
    \end{tabular*}
  \caption{mAP comparison to~\cite{Sun:ECCV:14} on the YouTube dataset.}
  \vspace*{-0.3cm}
  \label{tab:comparewithECCV}
\end{table}

\subsubsection{Comparison to Supervised Learning}  \vspace*{-0.2cm}
We have also compared our method to the latent ranking SVM technique in~\cite{Sun:ECCV:14}, using their YouTube dataset and their mAP evaluation metric. We note that our method is at a significant disadvantage in this comparison, as our system is trained using only the edited videos in the dataset, in contrast to~\cite{Sun:ECCV:14} which also utilizes the unedited counterparts. In addition, as the number of edited videos in this training set is relatively small, there is a risk of over-fitting as our system is primarily designed to leverage large-scale web data. The results, listed in Table~\ref{tab:comparewithECCV}, show that even though the supervised method benefits from major advantages in this comparison, the performance gap on this testing set is small for \textit{dog, gym, parkour} and \textit{surfing}. Moreover, as many of the training and testing video clips are from the same raw videos, it is particularly hard in this case for an unsupervised method such as ours to obtain good results relative to a supervised method trained on pre- and post-edit pairs.

We also examined latent ranking SVM trained on their own data with C3D features, but applied to our testing set. The results are shown in first column of Table~\ref{tab:overall_results} on domain categories that are shared by the two datasets. It can be seen that LRSVM does not perform as well as our RRAE. A possible explanation is that the supervised learning has a high risk of overfitting due to the limited training data, and as a result it may not generalize well to large-scale testing sets.

\vspace*{-0.1cm}
\section{Conclusion}\vspace*{-0.2cm}
We presented a scalable highlight extraction method based on unsupervised learning. Our technique relies on an improved auto-encoder with two significant modifications: a novel shrinking exponential loss which reduces sensitivity to noisy training data crawled from the web, and a recurrent auto-encoder configuration with LSTM cells. Generalizing this technique to other video processing problems would be a potential avenue for future work.

{\small
\bibliographystyle{ieee}
\bibliography{egbib}
}

\end{document}